\documentclass[10pt,twocolumn,letterpaper]{article}
\usepackage{times}
\usepackage{authblk} 

\usepackage{cvpr}              % To produce the CAMERA-READY version

\usepackage[dvipsnames]{xcolor}
\newcommand{\cxb}[1]{\textcolor{black}{#1}}
\newcommand{\cx}[1]{\textcolor{black}{#1}}
\newcommand{\lx}[1]{\textcolor{black}{#1}}

\graphicspath{{CVPR25-LX/images/}}

\usepackage{algpseudocode}
\usepackage{algorithm}

\usepackage{pifont}
\usepackage{multirow}

\definecolor{cvprblue}{rgb}{0.21,0.49,0.74}
\usepackage[pagebackref,breaklinks,colorlinks,allcolors=cvprblue]{hyperref}
\usepackage[capitalize]{cleveref}

\def\paperID{4838} % *** Enter the Paper ID here
\def\confName{CVPR}
\def\confYear{2025}

\title{LoRA Subtraction for Drift-Resistant Space in Exemplar-Free Continual Learning}

\author[1]{\vspace{-0.4cm}Xuan Liu}
\author[1,2,3]{Xiaobin Chang\thanks{indicates corresponding author.}}
\affil[1]{School of Artificial Intelligence, Sun Yat-sen University, China}
\affil[2]{Key Laboratory of Intelligent Assessment Technology for Sustainable Tourism, Ministry of Culture and Tourism, Sun Yat-sen University}
\affil[3]{Key Laboratory of Machine Intelligence and Advanced Computing, Ministry of Education, China}
\affil[ ]{\ttfamily\small liux687@mail2.sysu.edu.cn, changxb3@mail.sysu.edu.cn}
\begin{document}
\maketitle
\begin{abstract}
In continual learning (CL), catastrophic forgetting often arises due to feature drift.
This challenge is particularly prominent in the exemplar-free continual learning (EFCL) setting, where samples from previous tasks cannot be retained, \lx{making it difficult to preserve prior knowledge.}
% Therefore, the model struggles to maintain prior knowledge, leading to a more significant performance drop on an older task.
\lx{To address this issue, some EFCL methods aim to identify feature spaces that minimize the impact on previous tasks while accommodating new ones.}
% To ensure consistent representations across tasks, it is vital to mitigate feature drift.
% Some EFCL methods aim to identify feature spaces that minimize the impact on previous tasks while accommodating new ones.
However, they rely on static features or outdated statistics stored from old tasks, which prevents them from capturing the dynamic evolution of the feature space in CL, leading to performance degradation over time.
In this paper, we introduce the Drift-Resistant Space (DRS), which effectively handles feature drifts without requiring explicit feature modeling or the storage of previous tasks.
A novel parameter-efficient fine-tuning approach called Low-Rank Adaptation Subtraction (LoRA$^-$) is proposed to develop the DRS.
This method subtracts the LoRA weights of old tasks from the initial pre-trained weight before processing new task data to establish the DRS for model training.
Therefore, LoRA$^-$ enhances stability, improves efficiency, and simplifies implementation.
Furthermore, stabilizing feature drifts allows for better plasticity by learning with a triplet loss.
\lx{Our method consistently achieves state-of-the-art results, especially for long task sequences, across multiple datasets.\footnote{Code is available at \url{https://github.com/scarlet0703/LoRA-Sub-DRS}.}}
% Extensive experiments across multiple datasets show that our method consistently achieves state-of-the-art results, particularly for long sequences of learning tasks.
% The code is publicly available\footnote{\url{https://github.com/scarlet0703/LoRA-Sub-DRS}}.
\end{abstract}
\section{Introduction} \label{sec:intro}
\begin{figure}[t]
  \centering
   \includegraphics[width=\linewidth]{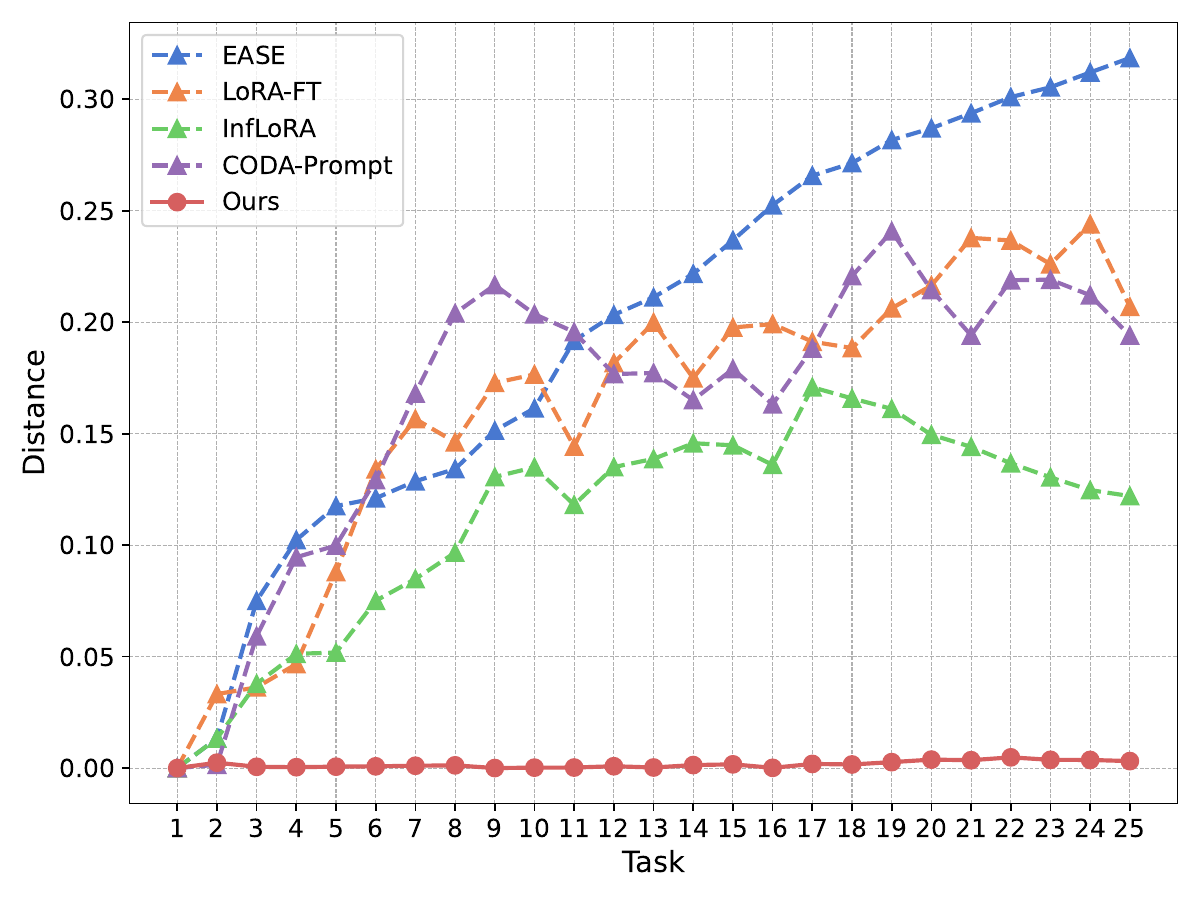}
   \caption{Illustrating the old task feature drifts of different EFCL methods. Imagenet-R dataset under 25 incremental tasks are used.}
   \label{fig:feat_drifts}
\end{figure}

% Please follow the steps outlined below when submitting your manuscript to the IEEE Computer Society Press.
% This style guide now has several important modifications (for example, you are no longer warned against the use of sticky tape to attach your artwork to the paper), so all authors should read this new version.
% cxb: abstract是骨架，intro按着骨架进行拓展，Intro的逻辑和关键词需要与abstract一致。把你想到的内容按逻辑写上去。
% 1: 介绍background：CL and our focus setting: EFCL, challenging在哪里？将灾难性遗忘和特征漂移联系起来：没有exemplar，特征漂移prones to happen，更加容易引发灾难性遗忘问题。
Continual learning (CL) methods~\cite{parisi2019continual,de2021continual} aim to enable deep models to acquire new knowledge continually, such as recognizing new object categories, while handling catastrophic forgetting~\cite{kemker2018measuring,mccloskey1989catastrophic} of previously learned information.
% Rehearsal-based methods~\cite{bonicelli2022effectiveness, yoon2021online, rebuffi2017icarl,castro2018end,douillard2020podnet} can help mitigate the forgetting problem by storing a few exemplars from previous tasks in memory and replaying them alongside new task data during training. However, access to these old task exemplars can be restricted due to data privacy concerns or memory capacity limitations.
% 
% Therefore, exemplar-free continual learning (EFCL) settings~\cite{goswami2024fecam,malepathirana2023napa,petit2023fetril,yu2020semantic} pose significant challenges, as managing feature drift from old tasks becomes increasingly difficult, which exacerbates catastrophic forgetting.
\lx{Rehearsal-based methods~\cite{bonicelli2022effectiveness, yoon2021online, rebuffi2017icarl,castro2018end,douillard2020podnet} alleviate forgetting by storing a small subset of past task exemplars and replaying them alongside new data during training.} However, access to such exemplars is often restricted due to privacy concerns or memory limitations.
\cx{Therefore, the study of exemplar-free continual learning (EFCL) becomes necessary.
This setting is challenging as no sample of previous tasks can be retained.}
% \lx{necessitating the development of exemplar-free continual learning (EFCL) strategies. EFCL settings~\cite{goswami2024fecam,malepathirana2023napa,petit2023fetril,yu2020semantic} introduce additional challenges, as models must retain past knowledge without direct memory access to previous data. 
\lx{Without rehearsal, feature representations of old tasks drift over time, leading to severe catastrophic forgetting, as shown in~\cref{fig:feat_drifts}.}

% Real-world applications often involve a continual influx of new data classes~\cite{gomes2017survey}, requiring models to adapt incrementally.
% To address this, continual learning (CL) has emerged as a solution, enabling models to learn from data streams while preserving knowledge acquired from earlier tasks.
% This poses a significant challenge known as catastrophic forgetting~\cite{kemker2018measuring,mccloskey1989catastrophic}, where models experience performance degradation on previously learned tasks after being fine-tuned on new ones.
% In exemplar-free CL (EFCL) settings~\cite{goswami2024fecam,malepathirana2023napa,petit2023fetril,yu2020semantic} where storing previous task samples is prohibited, most CL methods fail to mitigate feature drift.
% This, in turn, exacerbates the problem of catastrophic forgetting, as the model cannot accurately restore the distribution of old class samples in the newly adjusted feature space. Therefore, it is crucial to counteract this drift to enhance model stability.
% 2. 现有方法介绍【PEFT为主】，good side，
% 从architecture-based methods开始，引入PEFT。
% PEFT: 引入预训练模型，为了充分利用其中的知识，freezing提升stability。为了提升plasticity，同时提升效率，现有方法常常使用PEFT，例如Prompt和Lora。【实验上应该加入prompt类的方法进行对比】。
% 最后指出相关问题、不足？？
\cxb{
Conventional architecture-based \cx{EFCL} methods~\cite{chen2023dynamic,douillard2022dytox,wang2022foster,yan2021dynamically} involve freezing certain parameters that were trained for the initial task while either adjusting other parameters or expanding the backbone for subsequent tasks.
However, they require substantial memory consumption and significant computational resources.
% However, these methods often require a considerable amount of memory to store exemplars and significant computational resources to fine-tune on new tasks.
% prompt方法为主？
% \lx{To address these,} 
\cx{To handle this,} 
we adopt recent strategies~\cite{liang2024inflora} that utilize parameter-efficient fine-tuning (PEFT)~\cite{wang2022learning,smith2023coda,gao2023unified} to enhance both the effectiveness and efficiency of the EFCL model.
Specifically, our model is built on a frozen backbone pre-trained on a large-scale isolated dataset, providing extensive general knowledge.
Moreover, low-rank adaptation (LoRA)~\cite{hu2021lora} is integrated to support efficient learning for downstream tasks.}

\cxb{
% prompt需要加入图中吗？
Incorporating the PEFT techniques, e.g., LoRA~\cite{hu2021lora} and prompt tuning~\cite{lester2021power}, in EFCL methods alone may not be sufficient to handle catastrophic forgetting. This is demonstrated by the increasing feature drifts shown in \cref{fig:feat_drifts} and the inferior results displayed in \cref{tab:feat_drifts}.
% Some Existing method sum/ideas
To mitigate feature drifts, existing methods~\cite{liang2024inflora,wang2021training,kong2022balancing} create feature subspaces that aim at preventing the learning of new tasks from interfering with the old ones.
% 有效（表）不足（图）
Despite their improved overall performance indicated in \cref{tab:feat_drifts}, these methods still exhibit noticeable feature drifts, as shown by the upward trends in \cref{fig:feat_drifts}.
% 原因分析（而我们避免了这个问题：在EFCL设定下，对旧任务的特征空间进行精确建模用于构建interference free subspace / null space）
The key to addressing this issue lies in capturing the dynamic evolution of the feature space for old tasks in CL.
However, in the EFCL setting, this is challenging because only static features and statistics from old tasks are available, not to mention the additional memory required to store such data.
Therefore, the subspaces created by these methods do not effectively minimize interference from new task learning, especially regarding older tasks that are based on more outdated information.
% The effectiveness of these methods primarily relies on capturing the dynamic evolution of the feature space for old tasks in CL.
% Therefore, the subspaces created do not sufficiently reduce the interference from learning new tasks in relation to previous ones, especially regarding older tasks that are associated with more outdated data.
}

\begin{table}[t]
\centering
\caption{Average accuracy of PEFT methods on Imagenet-R dataset.}
\label{tab:feat_drifts}
\begin{tabular}{c|ccc}
\hline
% \hline
% Task              & 10    & 25    & 50    \\ \hline
\multirow{2}{*}{Method} & \multicolumn{3}{c}{Task}                                       \\ \cline{2-4} 
                        & \multicolumn{1}{c}{10}    & \multicolumn{1}{c}{25}    & 50    \\ \hline
LoRA-FT     & 76.03 & 67.13 & 58.56 \\ 
CODA-Prompt & 76.95 & 68.74 & 52.79 \\ 
InfLoRA     & 75.10 & 74.73 & 66.27 \\ 
EASE        & 81.27 & 79.32 & 75.35 \\ \hline
\textsc{LoRA$^-$DRS} (Ours)      & \textbf{81.52} &\textbf{ 79.79} & \textbf{77.93} \\ 
\hline
% \hline
\end{tabular}
\end{table}

\cxb{% In this work, we introduce a novel space for model training in the EFCL setting. As illustrated in \cref{fig:feat_drifts}, its curve remains consistently low and flat across different tasks, indicating that the feature drifts are effectively controlled. Therefore, the proposed space is called Drift-Resistant Space (DRS).
In this paper, we present Drift-Resistant Space (DRS) for model training in the EFCL setting. As illustrated in \cref{fig:feat_drifts}, our method maintains a consistently low and stable curve across tasks, indicating effective management of feature drifts. Therefore, this approach significantly mitigates catastrophic forgetting, as evidenced by its superior stability and overall performance in \cref{tab:feat_drifts}.
To avoid relying on static information from old tasks, we propose a novel approach called LoRA Subtraction (LoRA$^-$) for establishing the DRS.
Specifically, we first subtract the layer-wise LoRA weights associated with previous tasks from the initial pre-trained weights. The modified model then processes data from new tasks to obtain the projection matrix for the DRS.
This simple weight subtraction operation allows the model to "forget" or "unlearn" the knowledge of the corresponding tasks~\cite{ilharco2022editing}.
Consequently, training networks in DRS can focus more on learning new tasks while effectively reducing interference with older ones.}
\cxb{The contributions of this work are summarized as follows:
\begin{itemize}
  \item We introduce DRS, a new space for model training in the EFCL setting that effectively resolves feature drifts and enhances stability in continual learning.
  % . It is demonstrated that learning in DRS effectively addresses feature drifts, thus enhancing stability in continual learning.
  \item LoRA$^-$ efficiently establishes DRS without requiring explicit feature modeling or storing static old data. Instead, it aims to reduce the influence of previous tasks in the parameter space.
  % LoRA$^-$ is a novel PEFT technique that diminishes the influence of old task knowledge on new tasks in a simple yet effective manner. It establishes DRS efficiently without the need for explicit feature modeling or storage of previous tasks.
  \item Our method~\textsc{LoRA$^-$DRS} is less affected by feature drifts, allowing us to use a triplet loss based on prototypes from previous tasks to further enhance its plasticity.
  % Our method struggles less with feature drifts and thus enables a triplet loss based on prototypes from previous tasks to improve its plasticity further.
  % Our method struggles less with feature drifts and can better benefit from a triplet loss to improve its plasticity.
  % Using a simple triplet loss with prototypes from previous tasks enhances our method's plasticity by controlling feature drifts. In contrast, other approaches struggle with feature drifts, reducing the effectiveness of their triplet loss and thus with limited improvements.
  % Using a simple triplet loss based on prototypes from previous tasks, the plasticity of our method can be further improved due to the feature drifts being more controlled. Other approaches lack controlled feature drifts, limiting their triplet loss effectiveness and resulting in less improvement in plasticity.
  % With the more controlled feature drifts, the proposed method can boost plasticity further using a triplet loss based on prototypes from previous tasks.
  % Furthermore, stabilizing feature drifts allows for better plasticity by learning with a triplet loss.
  % plasticity. triplet loss, exclusive?
\end{itemize}}
\cxb{Extensive experiments are conducted in a challenging continual learning setting known as exemplar-free class incremental learning (EFCIL). The proposed method achieves state-of-the-art performance, particularly for long sequences of learning tasks across multiple datasets, highlighting its effectiveness and efficiency.}

\section{Related Work}
\label{sec:relatedwork}

% EFCIL侧重于讲这个任务是一个很难的EFCL设定（不是well established），突出难在哪里，tasks之间的类别不重合，而且没有exemplar，对model stability很不利。你要讲述的是现有的EFCIL方法怎么增强stability，总结出他们共同的不足，例如要存储旧类特征之类的。然后讲一下我们通过控制特征漂移保持stability，同时还能利用triplet loss增强了plasticity
% \subsection{Exemplar-Free Class-Incremental Learning (EFCIL)} 
\subsection{Exemplar-Free Class-Incremental Learning} 
% 应该是EFCL，不只是EFCIL。EFCIL是EFCL的一部分.
% 设定上，需要先回顾EFCL，给出总结。然后关注讨论EFCIL这个设定，突出challenging。
% 方法上，EFCIL的方法至少分成两部分，一部分是from scratch，另一部分是基于与训练模型（即2.2）
% 总体Related Work可以分两节，把2.2合并至2.1，2.2内容另起一段重点讨论即可。
Class-Incremental Learning (CIL) is a well-established paradigm in continual learning, designed to learn new classes while reducing catastrophic forgetting ~\cite{ahn2021ss,kim2021split,zhang2020knowledge}. Early Computer Vision methods preserved knowledge by storing raw data or features, but rising privacy concerns and storage limits have spurred interest in Exemplar-Free Class-Incremental Learning (EFCIL).
% Class-Incremental Learning (CIL) is a well-established paradigm in continual learning, aiming to continuously learn new classes while mitigating the issue of catastrophic forgetting in old classes~\cite{ahn2021ss,kim2021split,zhang2020knowledge}. In the domain of Computer Vision, early methods often relied on storing raw data or features from previous tasks to preserve knowledge. However, with increasing concerns over privacy and the storage demands associated with retaining raw samples, there has been a growing interest in Exemplar-Free Class-Incremental Learning (EFCIL).
One notable approach is LwF~\cite{dhar2019learning}, which mitigates forgetting by constraining the current model’s outputs to stay close to those of the previous model. PASS~\cite{zhu2021prototype} improves on this by incorporating self-supervised learning in the backbone, followed by functional regularization and feature rehearsal. SSRE~\cite{zhu2022self} proposes a novel architecture design that facilitates the transfer of invariant knowledge across tasks. In FeTRIL~\cite{petit2023fetril}, the authors freeze the feature extractor and leverage the variance in the current task’s data to estimate the location of old class features. FeCAM~\cite{goswami2024fecam} adopts a similar strategy by using the mean and covariance of previous task features, proposing a Mahalanobis distance-based classifier for better task separation.
In this paper, we adopt the EFCIL setting, where old data cannot be stored due to privacy concerns or device limitations, and aim to advance methods that enable continual learning without reliance on exemplars.

%-------------------------------------------------------------------------PEFT侧重于讲这些方法假设固定旧任务参数就能保证stability，没有关注特征漂移的问题，灾难性遗忘依旧存在，导致效果下降明显，特别是在long CL tasks设定下。
% \subsection{Parameter-Efficient Fine-Tuning (PEFT)}
\subsection{Parameter-Efficient Fine-Tuning}

% The advent of large-scale pre-trained models (PTMs) has sparked significant interest in fine-tuning techniques that effectively adapt these models to downstream tasks. Traditional continual learning (CL) methods typically involve training models from scratch, but recent advancements have shifted focus to CL with pre-trained models~\cite{wang2022s,wu2022class,zhou2023revisiting,zhou2022learning}, leveraging their strong feature representations.
The advent of large-scale pre-trained models (PTMs) has sparked significant interest in fine-tuning techniques for downstream tasks. 
Traditional continual learning (CL) trains models from scratch, but recent work explores CL with PTMs~\cite{wang2022s,wu2022class,zhou2023revisiting,zhou2022learning}, leveraging their strong feature representations.
% Traditional continual learning (CL) methods typically involve training models from scratch, but recent advancements have shifted focus to CL with PTMs~\cite{wang2022s,wu2022class,zhou2023revisiting,zhou2022learning}, leveraging their strong feature representations.
% Parameter-Efficient Fine-Tuning (PEFT) is a transfer learning approach that avoids full fine-tuning of a PTM. Instead, it involves inserting and fine-tuning specific sub-modules within the network, where the pre-trained parameters remain frozen. 
Parameter-Efficient Fine-Tuning (PEFT) avoids full PTM fine-tuning by inserting and tuning specific sub-modules while keeping pre-trained parameters frozen.
% This technique has demonstrated effective transfer learning results, particularly in Natural Language Processing (NLP)~\cite{houlsby2019parameter,li2021prefix}. 
Among the most widely used techniques in PEFT are LoRA~\cite{hu2021lora} and Prompt~\cite{lester2021power}, which have shown strong performance in CL~\cite{gao2023unified,smith2023coda,wang2022dualprompt}.
L2P~\cite{wang2022learning} builds on PTMs by learning additional dynamic prompts that guide the model to solve specific tasks. DualPrompt~\cite{wang2022dualprompt} introduces two separate prompt spaces—general and expert prompts—encoding task-invariant and task-specific instructions, respectively. CODAPrompt~\cite{smith2023coda}presents a decomposed attention-based continual learning method that offers greater capacity for learning than previous prompt-based approaches~\cite{wang2022dualprompt,wang2022learning}. 
In contrast, LAE~\cite{gao2023unified} and ADAM~\cite{zhou2023revisiting} introduce unified frameworks for PEFT through model ensembling. SLCA~\cite{zhang2023slca} builds upon the Gaussian modeling of old task features proposed by~\cite{zhu2021prototype}, using it to adjust classifiers.
However, all these methods rely on fixing the parameters of previous tasks to maintain stability. While they fail to address the issue of feature drift, which prevents the model from maintaining good performance and avoiding forgetting over long-term tasks.

%------------------------------------------------------------------------- Mitigating Feature Drift侧重于讲各个方法是通过找到特定的空间进行模型学习，以图降低新任务学习就旧任务特征的影响。举例分析说明。总结，这些方法都依赖于旧数据，是static的，不能对dynamic的CL过程精确建模，因而效果受限。而且实现上还需要存储旧数据。概括说明我们的方法避免了这些问题，控制了feature drifts，显著缓解了灾难性遗忘问题。
\subsection{Mitigating Feature Drift}
A critical aspect in EFCIL is the feature drift in old classes when new tasks are learned without old samples.
Some approaches aim to correct feature drift. 
For example, SDC~\cite{yu2020semantic} and ~\cite{toldo2022bring} estimate and compensate for feature drift with current-task samples after each incremental phase. Methods like NAPA-VQ~\cite{malepathirana2023napa} and Prototype Reminiscence~\cite{shi2023prototype} enhance or reshape old prototypes using topological information or interpolation with new samples. ADC~\cite{goswami2024resurrecting} generates adversarial samples as pseudo-exemplars to measure drift. FeTrIL~\cite{petit2023fetril} assumes similar feature distributions across all classes and use prototypes as the nearest feature center of new classes. EASE~\cite{zhou2024expandable} designs a semantic-guided prototype complement strategy that recalculates prototypes in the new feature space without relying on old exemplars.
Other approaches focus on minimize the interference of the new task on the old tasks. InfLoRA~\cite{liang2024inflora} uses gradient information from old tasks to design a subspace for LoRA’s dimensionality reduction matrix, reducing the impact of new tasks on old ones. Adam-NSCL~\cite{wang2021training} and AdNS~\cite{kong2022balancing} optimizes network parameters by using input features from previous tasks to obtain the approximate null space of old tasks.
However, despite the impressive performance achieved by these methods, several challenges remain in effectively controlling feature drift. These approaches assume that the model learns new concepts in an interference-free space of previous tasks. Since this space is computed based on stored, static statistics of old tasks, it becomes difficult to accurately model the highly dynamic feature space in continual learning. As the number of tasks increases, those statistics become increasingly outdated, leading to a noticeable decline in the performance of these methods.
\section{Methodology}
\label{sec:method}
% two stages
\begin{figure*}[t]
  \centering
   % \includegraphics[width=0.9\linewidth]{pipeline.pdf}
   \includegraphics[width=0.8\linewidth]{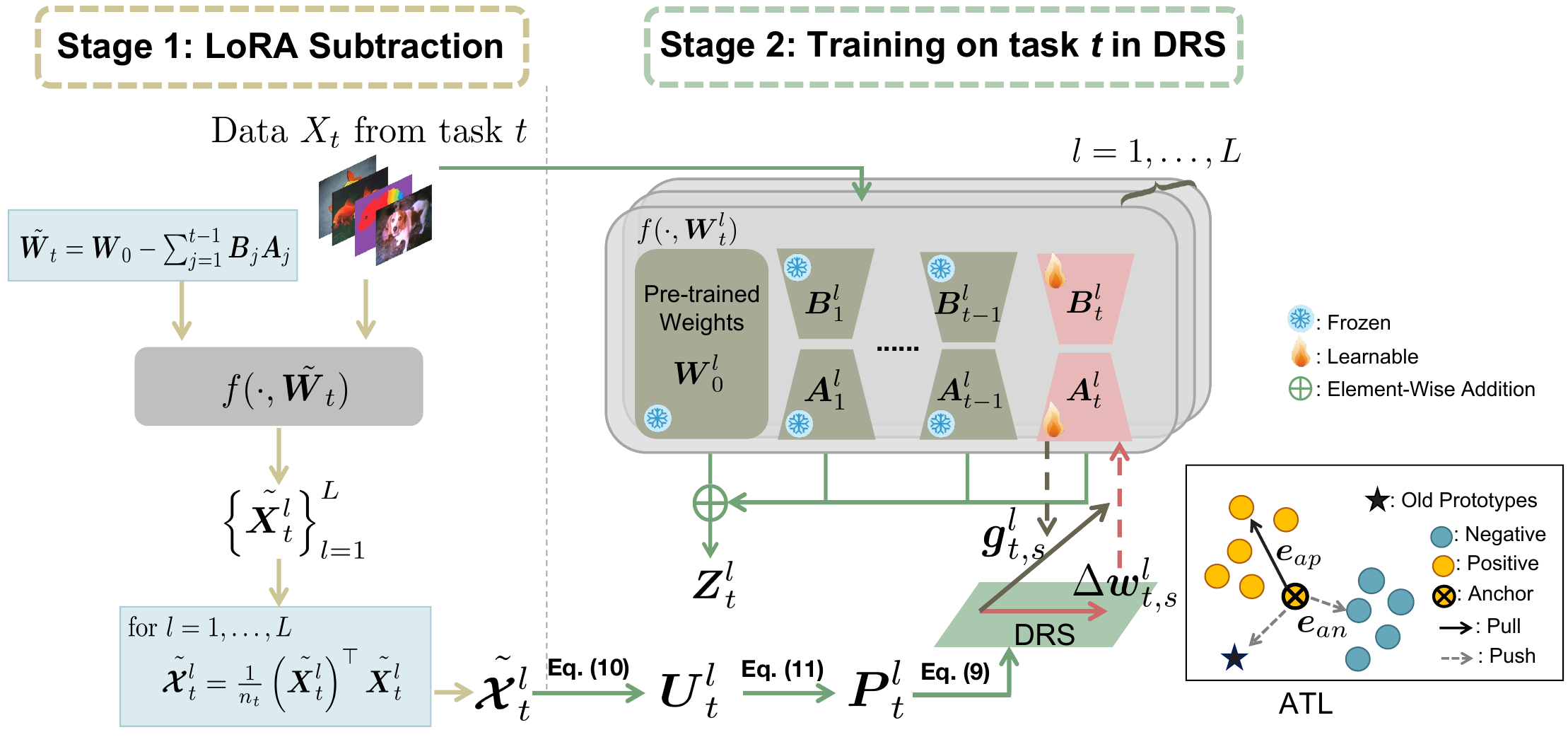}
   \caption{The training pipeline of the proposed LoRA Subtraction for Drift-Resistant Space. \lx{Before training on the \( t \)-th task, LoRA subtraction is applied to construct the drift-resistant space (DRS).
   % using \( X_t \). 
   During training, pre-trained weights and previously learned \cx{LoRAs are frozen.}
   % branches remain frozen. 
   \( A_t \) and \( B_t \) of \cx{the current task is learned} by projecting gradients into DRS, with augmented triplet loss (ATL) enhancing plasticity.
   % class separation and learning plasticity.
   }}
   \label{fig:drs}
\end{figure*}

\subsection{Preliminary}
%In EFCIL, we have a sequence of tasks \( \{1, \dots, T\} \) where new classes emerge over time, and we are not allowed to store samples from old classes. During the training of task \( t \), we have access only to the current dataset \( D_t = \{X_t, Y_t\} \), where \( X_t \) represents the images and \( Y_t \) represents the labels. The goal of EFCIL is to sequentially train a model \( f \) with parameters \( \boldsymbol{W} \) on each task, ensuring that it maintains good performance across all tasks.
Exemplar-free class-incremental learning (EFCIL) is designed to address the challenges of continual learning without exemplars, where the model must learn from a data stream of \( T \) sequential tasks, indexed as \( \{1, 2, \dots, T\} \), with non-overlapping class distributions. For each task \( t \), the dataset is defined as \( D_t = \left\{ (x_{i, t}, y_{i, t}) \right\}_{i=1}^{n_t} \), where \( x_{i, t} \) is an input sample, \( y_{i, t} \) is its label, and \( n_t \) is the number of instances in task \( t \). An instance \( x_{i, t} \) belongs to class \( y_{i, t} \), where \( y_{i, t} \in Y_t \) and \( Y_t \) is the label space of task \( t \), and \( Y_t \cap Y_{t'} = \emptyset \) for \( t \neq t' \), i.e., non-overlapping classes for different tasks. 
During training on task \( t \), only the dataset \( D_t \) is accessible, and no samples from previous tasks can be stored or used. 

Following the framework of parameter-efficient fine-tuning (PEFT) for continual learning~\cite{wang2022dualprompt, wang2024hierarchical} , we assume the model \( \mathcal{M} _{\boldsymbol{\theta}} \) is a pre-trained model with weights \( \boldsymbol{\theta} \).
Instead of updating the model parameters directly, Low-Rank Adaptation (LoRA) incorporates auxiliary low-rank matrices \( \boldsymbol{B} \) and \( \boldsymbol{A}\) into the model’s linear layers, while keeping the pre-trained weight matrix in each self-attention block \( \boldsymbol{W}_0\) fixed. Assume \(f(\cdot, \boldsymbol{W})\) denotes the linear transformation layer with parameters \(\boldsymbol{W}\in \mathbb{R}^{d \times d} \).
%(such as the \( \boldsymbol{Q} \), \( \boldsymbol{K} \), \( \boldsymbol{V} \), and \( \boldsymbol{O} \) matrices in a self-attention block~\cite{vaswani2017attention}), while keeping the pre-trained weight matrix \( \boldsymbol{W}_0 \in \mathbb{R}^{d \times k} \) fixed. 

Our approach consists of two main stages. In the first stage, before learning the \( t \)-th new task, the parameters \(\boldsymbol{W}_{t-1}\) are frozen, and a LoRA branch \(\Delta \boldsymbol{W}_{t} = \boldsymbol{B}_{t} \boldsymbol{A}_{t} \) is expanded, where \( \boldsymbol{A}_t \in \mathbb{R}^{r \times d} \) is a dimensionality reduction matrix and \( \boldsymbol{B}_t \in \mathbb{R}^{d \times r} \) is a dimensionality expansion matrix, with rank \( r \ll d \). 
To make the learning process resistant to feature drift, we apply LoRA Subtraction to obtain a drift-resistant space (DRS), which stabilizes the learning of new tasks while minimizing interference with previously learned knowledge.

In the second stage, we train the LoRA parameters \( \boldsymbol{A}_{t}  \) and \( \boldsymbol{B}_{t}  \) for the current task \( t \) in the DRS. Specifically, during training, when the data \( D_t \) from task \( t \) is fed to the model, the input and output features at the \( l \)-th linear layer within a self-attention block are denoted by \( \boldsymbol{X}_{t}^l\) and \( \boldsymbol{Z}_{t}^l\), respectively. The forward propagation in this layer can be represented by
\begin{equation}
\boldsymbol{Z}_{t}^l =  f(\boldsymbol{X}_{t}^l, \boldsymbol{W}_t^l), 
\label{eq:outputz}
\end{equation}
\begin{equation}
\boldsymbol{X}_{t}^{l+1} =  \sigma_l(\boldsymbol{Z}_{t}^l), 
\label{eq:inputx}
\end{equation}
\begin{equation}
\boldsymbol{W}_t^{l}  =   \boldsymbol{W}_{t-1}^{l} + \Delta \boldsymbol{W}_{t}^{l}    =\boldsymbol{W}_0 ^{l}  + \sum_{j=1}^{t}  \boldsymbol{B}_j^{l}  \boldsymbol{A}_j^{l}, 
\label{eq:wt}
\end{equation}
where \(\sigma\) is a nonlinear activation function, \( l = 1, \dots, L \), and \( \boldsymbol{X}_{t}^{1} = X_t\).
When training on task \( t \) at the \( s \)-th training step, the parameter update across the \( L \) linear layers is denoted as \(
\Delta \boldsymbol{w}_{t,s} = \{\Delta \boldsymbol{w}_{t,s}^1, \Delta \boldsymbol{w}_{t,s}^2, \dots, \Delta \boldsymbol{w}_{t,s}^L\}\). 

The complete pipeline of our method is illustrated in~\cref{fig:drs}.
% In summery, the DRS obtained in the first stage serves as the projection space in the second stage, ensuring the model remains resistant to feature drift across tasks. The method effectively preserves learned features while enabling efficient adaptation to new tasks. 
% For each incoming task \( t \), we first establish a drift-resistant space. This is achieved by performing LoRA Subtraction to obtain adjusted weights \( \tilde{\boldsymbol{W_t}} \), which, together with the current task’s data \( D_t \), are used to calculate the drift-resistant space \( S_t^l \). 
% Next, we train the LoRA parameters \( A_t \) and \( B_t \) for the current task \( t \). During each iteration, we compute the loss \( \mathcal{L}_{\text{total}} \) and calculate gradients with respect to \( D_t \). These gradients are then projected onto \( S_t^l \) before being applied to the model weights, effectively learning new information while preventing interference with prior tasks.

%According to the default initialization in LoRA~\cite{hu2021lora,mangrulkar2022peft}, it is common to initialize the matrix \( \boldsymbol{B} \) with an all-zero matrix and \( \boldsymbol{A} \) with Kaiming uniform~\cite{he2015delving} initialization. This technique significantly reduces the number of parameters that need to be fine-tuned, thereby decreasing computational and memory requirements at each step.

\begin{algorithm}[t]
\caption{LoRA Subtraction for Drift-Resistant Space}
\label{al:train}
\textbf{Inputs:} Datasets $D_t$ for task $ t \in \{1, 2, \dots , T \}$, a pre-trained ViT model $\mathcal{M} _{\boldsymbol{\theta}}$ with $L$ linear layers $f(\cdot, \boldsymbol{W})$, loss function $\mathcal{L}_{\text{total}}$, learning rate $\alpha$. \\
\textbf{Initialization:} Initialized parameters $ \boldsymbol{A}, \boldsymbol{B}, \boldsymbol{\theta}, \boldsymbol{W}$ .
\begin{algorithmic}[1]
% \State \texttt{\# continual learning tasks}
\For{task $t \in \{1, 2, \dots , T \}$}
    \If{$t > 1$}
        % \State \texttt{\# Stage 1: Compute the drift-resistant space}
        \State \texttt{\# Stage 1: Compute DRS}
        % \State Derive $\tilde{\boldsymbol{W_t}}$ through LoRA Subtraction using Eq.~(\ref{eq:wnt}).
        \State Compute $\tilde{\boldsymbol{\mathcal{X}}_{t}^{l}}$ through LoRA Subtraction with $D_t$ and $f(\cdot,\tilde{\boldsymbol{W}_t}^{l})$ as inputs using~\cref{eq:x}.
    \EndIf

    % \State \texttt{\# Stage 2: Train $ A_t $ and $ B_t $ in drift-resistant space}
    \State \texttt{\# Stage 2: Train $ A_t $ and $ B_t $ in DRS}
    \State Expand a LoRA branch $\Delta  \boldsymbol{W}_{t} = \boldsymbol{B}_{t} \boldsymbol{A}_{t}$.
    \State Compute $\boldsymbol{P}_t^l$ for each layer $l = 1, \dots, L$ using $\tilde{\boldsymbol{\mathcal{X}}_{t}^{l}}$ and~\cref{eq:space}.
    \State Set $s = 0$ and $\boldsymbol{W}_{t,0} = \boldsymbol{W}_{t-1} + \Delta  \boldsymbol{W}_{t, 0}$;
    \While{not converged}
        \State Sample a batch $\{X_t^b, Y_t^b\}$ from $D_t$.
        \State Compute loss $\mathcal{L}_{\text{total}}$ using $\{X_t^b, Y_t^b\}$ and $\boldsymbol{W}_{t,s}$ based on~\cref{eq:loss}.
        \State Get candidate parameter update $\boldsymbol{g}_{t,s} = \{\boldsymbol{g}_{t,s}^1, \dots, \boldsymbol{g}_{t,s}^L\}$ with respect to $\mathcal{L}_{\text{total}}$ with $\{x_t, y_t\}$ using Adam.
        \If{$t = 1$}
            \State $\Delta \boldsymbol{w}_{t,s}^l = \boldsymbol{g}_{t,s}^l, \quad l = 1, \dots, L$
        \Else
            \State $\Delta \boldsymbol{w}_{t,s}^l = \boldsymbol{P}_t^l \left(\boldsymbol{P}_t^l\right)^{\top} \boldsymbol{g}_{t,s}^l, \quad l = 1, \dots, L$
        \EndIf
        \State $\Delta  \boldsymbol{W}_{t,s+1}^l = \Delta  \boldsymbol{W}_{t,s}^l - \alpha \Delta \boldsymbol{w}_{t,s}^l, \quad l = 1, \dots, L$
        \State $s = s + 1$
    \EndWhile
\EndFor
\end{algorithmic}
\end{algorithm}

%-------------------------------------------------------------------------
\subsection{Stage 1: LoRA Subtraction}

The core objective of LoRA subtraction (LoRA$^-$) is to effectively remove information from previous tasks to create a drift-resistant space. 
Recent studies~\cite{ilharco2022editing} have shown that a task vector can direct specific changes in the weight space of a pre-trained model. 
Let \( \boldsymbol{\theta}_0 \) denote the weights of the pre-trained model and \( \boldsymbol{\theta}_t \) the weights after fine-tuning on task \( t \). The task vector for task \( t \) is given by
\begin{equation}
\boldsymbol{V}_t = \boldsymbol{\theta}_t - \boldsymbol{\theta}_0. 
\label{eq:t}
\end{equation}

By negating task vectors, performance on specific tasks can be reduced or forgotten with minimal impact on unrelated tasks. 

Building on this insight, LoRA$^-$ controls feature drift by selectively negating prior task influences.
After training on task \( t-1 \), the task vector \( \boldsymbol{V}_{t-1}^{l} \) for each linear layer \( l = 1, \dots, L \) is defined as
 % a summation over all previous tasks
\begin{equation}
\boldsymbol{V}_{t-1}^{l}  = \boldsymbol{W}_{t-1}^{l}  - \boldsymbol{W}_0^{l} = \sum_{j=1}^{t-1} \boldsymbol{B}_j^{l} \boldsymbol{A}_j^{l}.
\label{eq:vt}
\end{equation}

To achieve forgetting of specific old tasks, LoRA$^-$ applies a negated cumulative task vector \( -\boldsymbol{V}_{t-1} \) to the pre-trained weights before training on the new task \( t \) 
\begin{equation}
\tilde{\boldsymbol{W}_t}^{l} = \boldsymbol{W}_0^{l} - \boldsymbol{V}_{t-1}^{l} = \boldsymbol{W}_0^{l} - \sum_{j=1}^{t-1} \boldsymbol{B}_j^{l} \boldsymbol{A}_j^{l}.
\label{eq:wnt}
\end{equation}

We then define a drift-resistant space (DRS) for the new task, ensuring that learning task \( t \) does not affect the feature distribution of prior tasks. 
Existing works~\cite{rusu2016progressive,wang2021training,zeng2019continual} have shown that the gradient update in a linear layer lies within the span of the input data. Therefore, we leverage the input matrix of the new task \( t \) under \( \tilde{\boldsymbol{W}_{t}} \) to define a learning space that captures features specific to the new task without disturbing previous knowledge.

%Let \( \tilde{\boldsymbol{W}_{t} }= \{\tilde{\boldsymbol{w}_{t}^1}, \tilde{\boldsymbol{w}_{t}^2}, \dots, \tilde{\boldsymbol{w}_{t}^L}\} \) denote the LoRA$^-$ weights across each layer \( l = 1, \dots, L \). 

Before training on task \( t \), we feed \( D_t \) to the model with LoRA$^-$ linear layer \( f( X_t, \tilde{\boldsymbol{W}_{t}^{l}}) \) to obtain the input feature \( \tilde{\boldsymbol{X}_{t}^{l} }\) at each layer \( l \).
Following~\cite{wang2021training}, we compute the uncentered covariance matrix of these input features as follows
\begin{equation}
\tilde{\boldsymbol{\mathcal{X}}_{t}^{l}} = \frac{1}{n_{t}} \left( \tilde{\boldsymbol{X}_{t}^{l}} \right)^{\top} \tilde{\boldsymbol{X}_{t}^{l}},
\label{eq:x}
\end{equation}
where \( n_t \) is the number of samples in task \( t \), and \( \tilde{\boldsymbol{\mathcal{X}}_{t}^{l}}\) captures input learning space at layer \( l \) for task \( t \), playing a key role in defining DRS for stable task-specific learning.

% By applying Singular Value Decomposition (SVD) to \( \tilde{\boldsymbol{\mathcal{X}}_{t}^{l}} \), we have:
% \begin{equation}
% \boldsymbol{U}_t^l \boldsymbol{\Lambda}_t^l (\boldsymbol{U}_t^l)^{\top} = \text{SVD}(\tilde{\boldsymbol{\mathcal{X}}_{t}^{l}} )
% \label{eq:svd}
% \end{equation}

% where \(\boldsymbol{U}_t^l\) represents the directions of the principal components, \( \boldsymbol{\Lambda}_t^l  \) indicates the significance of each component, and \( (\boldsymbol{U}_t^l)^{\top}\) represents the transpose of \( \boldsymbol{U}_t^l\).

% To form the DRS for each layer \( l\), denoted as \( \boldsymbol{S_t^l }\), we select a subset of the singular vectors from \(\boldsymbol{ U^l} \) that correspond to the largest singular values, denoted as \( \boldsymbol{(U^l)_{p}} \). Inspired by Principal Component Analysis, this subset captures the most significant components of the input feature space in LoRA$^-$ network, ensuring a stable learning.
% Thus, \( \boldsymbol{S_t^l} \) is given by:
% \begin{equation}
% \boldsymbol{S_t^l} = \boldsymbol{(U^l)_{p}}
% \label{eq:space}
% \end{equation}

%-------------------------------------------------------------------------

% \subsection{Training Network in Drift-Resistant Space}
\subsection{Stage 2: Training in Drift-Resistant Space}

In this section, we introduce a network training algorithm designed for learning sequential tasks in a drift-resistant space (DRS). Our approach ensures that each new task can be learned with minimal interference from previously learned tasks, maintaining a stable feature representation. 

% When training on a new task \( t \), the weights are updated iteratively at each training step, with the parameters at step \( s \) denoted as \( \boldsymbol{W_{t,s}} = \{\boldsymbol{w_{t,s}^1}, \boldsymbol{w_{t,s}^2}, \dots, \boldsymbol{w_{t,s}^L}\} \). For task-specific adaptation, a LoRA branch is expanded as \(\boldsymbol{W_t} = \boldsymbol{W_{t-1}} + \boldsymbol{B_t A_t}\), where \( \boldsymbol{W_{t-1}} \) remains unchanged, \( \boldsymbol{B_t} \) and \( \boldsymbol{A_t} \) are learned specifically for task \( t \).
% Thus, the change in weights at each layer during step \( s \) on task \( t \) is represented by:
% \begin{equation}
% \Delta \boldsymbol{W}_{t,s}^{l}= \boldsymbol{B}_{t,s}^{l} \boldsymbol{A}_{t,s}^{l} = \{\Delta \boldsymbol{w}_{t,s}^1, \Delta \boldsymbol{w}_{t,s}^2, \dots, \Delta \boldsymbol{w}_{t,s}^L\}, 
% \label{eq:change}
% \end{equation}
% where \( \Delta \boldsymbol{W}_{t,s}^{l} \) corresponds to the task-specific LoRA adaptation applied at each layer \( l \) during step \( s \). 
When training on a new task \( t \), the LoRA weights \( \Delta \boldsymbol{W}_{t} \) are updated iteratively at each training step \( s \), with the parameter update at each linear layer represented as
\begin{equation}
\Delta \boldsymbol{w}_{t,s} = \{\Delta \boldsymbol{w}_{t,s}^1, \Delta \boldsymbol{w}_{t,s}^2, \dots, \Delta \boldsymbol{w}_{t,s}^L\}, 
\label{eq:change}
\end{equation}
where \( \Delta \boldsymbol{w}_{t,s}^{l} \) corresponds to the parameter updates in the LoRA branch for layer \( l \) during training step \( s \).

To mitigate feature drift, we project the gradients at each training step into a DRS before updating the weights. DRS is designed to preserve knowledge from previous tasks by restricting updates to directions that minimally interfere with learned representations. Specifically, for each layer \( l \), we denote the gradients at step \( s \) as \( \boldsymbol{g}_{t,s}^l \) and project them into DRS using a projection operator \( \boldsymbol{P}_t^l \). The projected update for each layer \( l \) is computed as follows
\begin{equation}
\Delta \boldsymbol{w}_{t,s}^l = \boldsymbol{P}_t^l \left( \boldsymbol{P}_t^l \right)^{\top} \boldsymbol{g}_{t,s}^l,
\label{eq:project}
\end{equation}
where \( \boldsymbol{P}_t^l \left( \boldsymbol{P}_t^l \right)^{\top} \) acts as a DRS projection operator.

Specifically, the DRS projection operator is obtained by applying Singular Value Decomposition (SVD) to the feature matrix \( \tilde{\boldsymbol{\mathcal{X}}_{t}^{l}} \) from LoRA subtraction (LoRA$^-$). Following methods in ~\cite{meyer2023matrix, wang2021training}, we apply SVD to \( \tilde{\boldsymbol{\mathcal{X}}_{t}^{l}} \)
\begin{equation}
\boldsymbol{U}_t^l \boldsymbol{\Lambda}_t^l (\boldsymbol{U}_t^l)^{\top} = \text{SVD}(\tilde{\boldsymbol{\mathcal{X}}_{t}^{l}}),
\label{eq:svd}
\end{equation}
where \( \boldsymbol{U}_t^l \) represents the directions of the principal components, \( \boldsymbol{\Lambda}_t^l \) indicates the significance of each component, and \( (\boldsymbol{U}_t^l)^{\top} \) is the transpose of \( \boldsymbol{U}_t^l \).

% To construct the DRS basis for layer \( l \), denoted as \( \boldsymbol{P_t^l }\), we select a subset of principal components \( (\boldsymbol{U}_t^l)_{k} \) from \( \boldsymbol{U}_t^l \) that correspond to the largest singular values. This selection criterion, inspired by Principal Component Analysis (PCA), captures the most significant directions in \( \tilde{\boldsymbol{\mathcal{X}}_{t}^{l}} \), where updates have minimal interference with previously learned knowledge. Thus, the DRS projection matrix for each layer \( l \) is defined as
% \begin{equation}
% \boldsymbol{P}_t^l = (\boldsymbol{U}_t^l)_{k},
% \label{eq:space}
% \end{equation}
% where \( (\boldsymbol{U}_t^l)_{k} \) includes only the top \( k \) singular vectors, selected based on 

% \begin{equation}
% % k = \min \left\{ j : \frac{\sum_{i=1}^{j} \lambda_i}{\sum_{i=1}^{n} \lambda_i} \geq \varepsilon  \right\},
% % \min_{k}   \left\{  \frac{\sum_{i=1}^{k} \lambda_i}{\sum_{i=1}^{d} \lambda_i} \geq \varepsilon  \right\},
% % \min_{k}   \frac{\sum_{i=1}^{k} \lambda_i}{\sum_{i=1}^{d} \lambda_i} \geq \varepsilon,
% \arg\min_{k}  \frac{\sum_{i=1}^{k} \lambda_i}{\sum_{i=1}^{d} \lambda_i}, s.t.ol{\Lambda}_t^l (\boldsymbol{U}_t^l)^{\top} = \text{SVD}(\tilde{\boldsymbol{\mathcal{X}}_{t}^{l}}),
% \label{eq:svd2}
% \end{equation}
% where \( \boldsymbol{U}_t^l \) represents the directions of the principal components, \( \boldsymbol{\Lambda}_t^l \) indicates the significance of each component, and \( (\boldsymbol{U}_t^l)^{\top} \) is the transpose of \( \boldsymbol{U}_t^l \).

To construct the DRS basis for layer \( l \), denoted as \( \boldsymbol{P_t^l }\), we select a subset of principal components \( (\boldsymbol{U}_t^l)_{k} \) from \( \boldsymbol{U}_t^l \) that correspond to the largest singular values. This selection criterion, inspired by Principal Component Analysis (PCA), captures the most significant directions in \( \tilde{\boldsymbol{\mathcal{X}}_{t}^{l}} \), where updates have minimal interference with previously learned knowledge. Thus, the DRS projection matrix for each layer \( l \) is defined as
\begin{equation}
\boldsymbol{P}_t^l = (\boldsymbol{U}_t^l)_{k},
\label{eq:space}
\end{equation}
where \( (\boldsymbol{U}_t^l)_{k} \) includes only the top \( k \) singular vectors, selected based on 
\begin{equation}
% k = \min \left\{ j : \frac{\sum_{i=1}^{j} \lambda_i}{\sum_{i=1}^{n} \lambda_i} \geq \varepsilon  \right\},
% \min_{k}   \left\{  \frac{\sum_{i=1}^{k} \lambda_i}{\sum_{i=1}^{d} \lambda_i} \geq \varepsilon  \right\},
\min_{k}   \frac{\sum_{i=1}^{k} \lambda_i}{\sum_{i=1}^{d} \lambda_i} \geq \varepsilon,
%%% revised
% \argmin_{k}  \frac{\sum_{i=1}^{k} \lambda_i}{\sum_{i=1}^{d} \lambda_i}, s.t. \frac{\sum_{i=1}^{k} \lambda_i}{\sum_{i=1}^{d} \lambda_i} \geq \varepsilon,
\label{eq:select}
\end{equation}
where \(\lambda_i\) represents the eigenvalues sorted in descending order, \(\varepsilon\) denote the cumulative variance threshold.

This DRS-based projection ensures that the updated weights remain in DRS, allowing the model to learn new tasks without degrading performance on prior ones.
% This ensures that DRS retains the most informative components while filtering out noisy or less relevant directions.

The detailed steps for both stages of our approach are outlined in~\cref{al:train}.

%-------------------------------------------------------------------------

\subsection{Model Optimization}
While drift-resistant space (DRS) controls feature drift to maintain stability in previously learned knowledge, we found that gradient projection can impair the model’s plasticity, hindering its ability to separate new classes in the feature space.

To address this, we propose an Augmented Triplet Loss (ATL), denoted as \( \mathcal{L}_{\text{TL}} \), which enhances class separation by maximizing the distance between positive and negative pairs for each anchor sample, thus improving plasticity. It is defined as
\begin{equation}
\mathcal{L}_{\text{TL}} = \max(0, e_{ap} - e_{an} + \epsilon ),
\label{eq:triplet}
\end{equation}
where \( e_{ap} \) and \( e_{an} \) are the hardest positive and hardest negative distances, and \( \epsilon \) is a margin parameter that enforces a desired separation between positive and negative pairs. 

% To compute \( e_{ap} \) and \( e_{an} \), we first calculate pairwise distances within the batch for the current task data \( D_t \) at task \( t \). Given an anchor sample \( x_i \) with label \( y_i \in Y^t \), where \( Y^t \) is the label set of the current task, we define the pairwise Euclidean distance matrix \( E \) as:
% \begin{equation}
% E_{i,j} = \| f(x_i) - f(x_j) \|_2
% \label{eq:euclidean}
% \end{equation}
% where \( f(x) \) denotes the feature embedding of sample \( x \). This matrix \( E \) provides the foundation for identifying the hardest positive and hardest negative samples for each anchor.

For an anchor sample \( x_{i, t} \) with label \( y_{i, t} \) in the \( t \)-th task, the hardest positive distance \( e_{ap} \) is the maximum distance between \( x_{i, t} \) and other samples with the same label
\begin{equation}
e_{ap} = \max_{j: y_{j, t} = y_{i, t}} E_{i,j},
\label{eq:ap}
\end{equation}
where \( E_{i,j} \) denotes the Euclidean distance metric between sample \( x_{i, t} \) and sample \( x_{j, t} \).

The hardest negative distance \( e_{an} \) considers both samples from the current task with labels \( y_{j, t}  \neq y_{i, t}  \) and prototypes \( p_c \in P_{1:t-1} \) from previous tasks, where \( P_{1:t-1} \) represents the prototype set of all previous tasks
\begin{equation}
e_{an} = \min\left( \min_{j: y_{j, t} \neq y_{i,t}} E_{i,j}, \min_{c} \| \mathcal{M} _{\boldsymbol{\theta}_t}(x_{i,t}) - p_{c} \|_2 \right),
\label{eq:an}
\end{equation}
where \( \mathcal{M} _{\boldsymbol{\theta}_t}(x_{i,t})\) represents the feature embedding of the anchor sample \( x_{i, t} \) under the current model parameters \( \boldsymbol{\theta}_t \), and \( \| \cdot \|_2 \) denotes the Euclidean distance to the prototype \( p_c \).

% The Cross-Entropy Loss \( \mathcal{L}_{\text{CE}} \) measures the difference between the predicted probability distribution and the true label distribution, defined as:
% \begin{equation}
% \mathcal{L}_{\text{CE}} = -\sum_{c=1}^{C} y_c \log(\hat{y}_c)
% \label{eq:ce}
% \end{equation}
% where \( C \) is the total number of classes, \( y_c \) is the true label indicator, and \( \hat{y}_c \) is the predicted probability for class \( c \).

The total loss \( \mathcal{L}_{\text{total}} \) combines cross-entropy loss \( \mathcal{L}_{\text{CE}}\) and the Augmented Triplet Loss, balanced by a weighting factor \( \lambda \)
\begin{equation}
\mathcal{L}_{\text{total}} = \mathcal{L}_{\text{CE}} + \lambda \mathcal{L}_{\text{TL}}.
\label{eq:loss}
\end{equation}

%-------------------------------------------------------------------------
\section{Experiments}
\label{sec:experiment}

\subsection{Experimental Setting}

\textbf{Datasets}\quad
We evaluate our method on multiple CIL benchmarks that are widely used in PEFT research. ImageNet-R~\cite{hendrycks2021many} is created by applying artistic transformations to 200 classes from the original ImageNet dataset~\cite{deng2009imagenet}. This dataset was introduced into the continual learning setting by prior work~\cite{wang2022dualprompt} and has since become a standard benchmark for evaluating continual learning methods based on PEFT. CIFAR100~\cite{krizhevsky2009learning}, consisting of 100 classes of small-scale images, is a commonly used benchmark dataset in CIL. 
% DomainNet~\cite{peng2019moment} consists of 345 classes from six diverse domains, offering a challenging benchmark due to its variability in both class distribution and visual domains. To address class imbalance, prior work~\cite{wang2023isolation} select the 200 categories with the most images, distributing them across tasks, each containing images from multiple domains.

\noindent\textbf{Implementation details}\quad
The proposed method is implemented in PyTorch~\cite{paszke2019pytorch}, following similar approaches to previous works~\cite{wang2022dualprompt, wang2022learning, smith2023coda, wang2024hierarchical, liang2024inflora}. Specifically, we utilize a pre-trained ViT-B/16-IN21K backbone and train the model with the Adam optimizer, using parameters $\beta_1 = 0.9$ and $\beta_2 = 0.999$. Training is conducted with a batch size of 128, across different datasets: each task is trained for 50 epochs on ImageNet-R, and 20 epochs on CIFAR100. Input images are resized to 224 × 224 and normalized within the range [0, 1]. Following previous work~\cite{gao2023unified, kumari2023multi, liang2024inflora}, we integrate the LoRA architecture into the key and value components of the ViT attention module.
\cx{Averaged performance of 5 runs is reported in the main results.}
% \lx{Reported metrics are averaged over five independent runs with different random seeds.} 
% we integrate the LoRA architecture by inserting it into the key and value components of the attention module.
% Training is conducted with a batch size of 128, across different datasets: each task is trained for 50 epochs on ImageNet-R, 20 epochs on CIFAR100, and 10 epochs on DomainNet. Input images are resized to 224 × 224 and normalized within the range [0, 1]. Following previous work~\cite{gao2023unified, kumari2023multi, liang2024inflora}, we integrate the LoRA architecture by inserting it into the key and value components of the attention module.

\noindent\textbf{Compared methods}\quad
 We compare against PTM-based methods specifically designed to mitigate feature drift, such as EASE~\cite{zhou2024expandable}, InfLoRA~\cite{liang2024inflora}, and Adam-NSCL~\cite{wang2021training}. For Adam-NSCL, we implement it with LoRA on PTM. In addition, we select several representative PEFT-based EFCIL methods for comparison, including \lx{LAE~\cite{gao2023unified},} L2P~\cite{wang2022learning}, DualPrompt~\cite{wang2022dualprompt}, and CODA-Prompt~\cite{smith2023coda}. We also establish a baseline, denoted as LoRA-FT, which sequentially finetunes the PTM using LoRA. For all methods, their best performance results are reproduced under our experimental settings to ensure fair and direct comparison.
 % We consider ViT-B/16-IN21K as the pre-trained backbone.
% We consider various pre-training paradigms as described in~\cite{wang2024hierarchical}, specifically Sup-21K, iBOT-1K, and DINO-1K, each trained on either ImageNet-21K or ImageNet-1K. 
% ConvPrompt~\cite{roy2024convolutional}
% C-LoRA~\cite{smith2023continual}
% HiDe-Prompt~\cite{wang2024hierarchical}

\noindent\textbf{Evaluation metrics}\quad
Following established practices in continual learning~\cite{zhou2024expandable, gao2023unified, lopez2017gradient}, we evaluate the model's performance through three key metrics:  accuracy after the last stage($ACC$), average accuracy ($\overline{ACC} $) and backward transfer ($BWT$). 
For accuracy, we denote $ACC_t$ as the model’s accuracy after completing the $t$-th task. To capture the overall performance, we use $ACC_T$, which represents the accuracy after the final task, and the average accuracy $\overline{ACC_T}$ across all tasks, defined as:
\begin{equation}
\overline{ACC_T} = \frac{1}{T} \sum_{t=1}^{T} ACC_t ,
\label{eq:acc}
\end{equation}
Backward transfer value assesses the effect of learning new tasks on previously learned tasks, capturing the average impact on prior knowledge retention in continual learning. A negative BWT value indicates forgetting, as the model's performance on prior tasks deteriorates when new tasks are added. We compute BWT as:
\begin{equation}
\text{BWT} = \frac{1}{T-1} \sum_{i=1}^{T-1} \left( A_{T,i} - A_{i,i} \right),
\label{eq:bwt}
\end{equation}
where $A_{T,i}$ denotes the accuracy on the $i$-th task after training on the final task $T$.

%-------------------------------------------------------------------------
% \subsection{Experimental Results}

\subsection{Main Results}
\begin{table*}[t]
\centering
% \caption{Comparison of different methods on ImageNet-R across various tasks.}
\caption{Averaged results \cx{(mean $\pm$ standard deviation)} of different methods trained under five random seeds of ImageNet-R.}
\label{tab:ming-IN21K}
\scalebox{0.85}{
\begin{tabular}{l|cc|cc|cc|cc}
\hline
\multicolumn{1}{c|}{Task} & \multicolumn{2}{c|}{10} & \multicolumn{2}{c|}{20} & \multicolumn{2}{c|}{25} & \multicolumn{2}{c}{50} \\ 
\hline
\multicolumn{1}{l|}{\rule{0pt}{2.5ex}Method} & \multicolumn{1}{c}{$ACC_{10}$} & \multicolumn{1}{c|}{$\overline{ACC}_{10}$} & \multicolumn{1}{c}{$ACC_{20}$} & \multicolumn{1}{c|}{$\overline{ACC}_{20}$} & \multicolumn{1}{c}{$ACC_{25}$} & \multicolumn{1}{c|}{$\overline{ACC}_{25}$} & \multicolumn{1}{c}{$ACC_{50}$} & $\overline{ACC}_{50}$ \\ 
\hline
LoRA-FT & 74.54$\pm$6.02& 73.43$\pm$4.52& 60.71$\pm$2.39& 72.53$\pm$1.72&  56.80$\pm$3.06& 70.06$\pm$1.74& 44.89$\pm$2.29& 59.25$\pm$1.40\\
LAE~\cite{gao2023unified} & 72.56$\pm$0.98& 78.56$\pm$0.47& 60.53$\pm$1.73& 76.64$\pm$0.91& 68.05$\pm$1.88& 75.40$\pm$1.31& 62.83$\pm$2.17& 70.83$\pm$1.63\\
L2P~\cite{wang2022learning} & 64.94$\pm$0.90& 70.33$\pm$1.90& 62.15$\pm$1.17& 68.35$\pm$2.12& 60.85$\pm$1.11& 67.17$\pm$2.46& 55.89$\pm$1.59& 62.98$\pm$2.89\\
DualPrompt~\cite{wang2022dualprompt} & 62.24$\pm$0.29& 74.63$\pm$0.94& 66.89$\pm$0.40& 73.07$\pm$1.21& 65.76$\pm$0.67& 72.15$\pm$1.19& 61.50$\pm$0.86& 68.63$\pm$1.31\\
CODA-Prompt~\cite{smith2023coda} & 72.15$\pm$0.12& 77.51$\pm$0.75& 67.53$\pm$0.24& 73.64$\pm$0.95& 63.86$\pm$0.81& 70.47$\pm$1.72& 48.89$\pm$0.90& 55.59$\pm$2.67\\
InfLoRA~\cite{liang2024inflora} & 74.95$\pm$0.90& 80.99$\pm$0.89& 71.46$\pm$0.95& 78.32$\pm$1.22& 69.09$\pm$0.93& 76.84$\pm$1.32& 60.49$\pm$1.43& 69.95$\pm$2.18\\
Adam-NSCL~\cite{wang2021training} & 72.24$\pm$0.69& 78.85$\pm$1.07& 65.52$\pm$1.11& 72.79$\pm$1.39& 62.04$\pm$1.74& 69.69$\pm$1.81& 49.82$\pm$2.67& 58.49$\pm$3.85\\
EASE~\cite{zhou2024expandable} & \textbf{75.94}$\pm$0.46& \textbf{81.67}$\pm$0.32& 73.78$\pm$0.44& 80.29$\pm$0.72& 72.69$\pm$0.39& 79.65$\pm$0.65& 68.54$\pm$0.71& 75.77$\pm$0.58\\ \hline
\textsc{LoRA$^-$DRS} (Ours) & 74.74$\pm$0.78& 81.16$\pm$0.59 & \textbf{74.80}$\pm$0.73& \textbf{80.69}$\pm$0.75& \textbf{74.19}$\pm$0.46& \textbf{80.06}$\pm$0.76& \textbf{72.12}$\pm$0.87& \textbf{77.94}$\pm$0.74\\ 
\hline
\end{tabular}}
\end{table*}

\begin{table*}[t]
\centering
% \caption{Comparison of different methods on CIFAR-100 across various tasks.}
\caption{Averaged results \cx{(mean $\pm$ standard deviation)} of different methods trained under five random seeds of CIFAR-100.}
\label{tab:cifar-IN21K}
\scalebox{0.85}{
\begin{tabular}{l|cc|cc|cc|cc}
\hline
\multicolumn{1}{c|}{Task} & \multicolumn{2}{c|}{10} & \multicolumn{2}{c|}{20} & \multicolumn{2}{c|}{25} & \multicolumn{2}{c}{50} \\ 
\hline
\multicolumn{1}{l|}{\rule{0pt}{2.5ex}Method} & \multicolumn{1}{c}{$ACC_{10}$} & \multicolumn{1}{c|}{$\overline{ACC}_{10}$} & \multicolumn{1}{c}{$ACC_{20}$} & \multicolumn{1}{c|}{$\overline{ACC}_{20}$} & \multicolumn{1}{c}{$ACC_{25}$} & \multicolumn{1}{c|}{$\overline{ACC}_{25}$} & \multicolumn{1}{c}{$ACC_{50}$} & $\overline{ACC}_{50}$ \\ 
\hline
LoRA-FT & 82.43$\pm$1.47& 89.30$\pm$1.00& 74.51$\pm$1.58& 84.10$\pm$0.95& 68.33$\pm$3.05& 80.81$\pm$1.02& 43.77$\pm$5.06& 63.45$\pm$2.90\\
LAE~\cite{gao2023unified} & 84.99$\pm$0.75& 89.36$\pm$0.84& 83.51$\pm$0.36&  88.15$\pm$0.59&  82.20$\pm$0.58&  87.02$\pm$0.87&  77.68$\pm$2.21& 83.16$\pm$1.10 \\
L2P~\cite{wang2022learning} & 83.41$\pm$0.60& 88.98$\pm$0.31& 80.72$\pm$1.12& 87.18$\pm$0.83& 79.54$\pm$0.72& 86.69$\pm$0.65& 73.91$\pm$1.67& 81.90$\pm$0.98\\
DualPrompt~\cite{wang2022dualprompt} & 86.40$\pm$0.62& 91.26$\pm$1.29& 83.82$\pm$0.51& 90.22$\pm$0.68& 82.65$\pm$1.28& 89.44$\pm$1.04& 76.66$\pm$0.74& 85.18$\pm$0.92\\
CODA-Prompt~\cite{smith2023coda} & 87.02$\pm$0.17& 91.39$\pm$0.23& 81.19$\pm$0.31& 87.27$\pm$0.35& 77.18$\pm$0.49& 84.56$\pm$0.31& 55.45$\pm$0.48& 68.39$\pm$0.53\\
InfLoRA~\cite{liang2024inflora} & 86.44$\pm$0.81& 91.16$\pm$0.79& 82.19$\pm$1.33& 88.05$\pm$0.64& 77.51$\pm$1.19& 85.02$\pm$1.61& 56.65$\pm$5.55& 70.29$\pm$1.86\\
Adam-NSCL~\cite{wang2021training} & 85.39$\pm$0.75& 89.67$\pm$0.53& 79.69$\pm$1.49& 85.52$\pm$1.10& 75.19$\pm$1.61& 81.29$\pm$1.90& 55.02$\pm$1.52& 66.27$\pm$1.34\\
EASE~\cite{zhou2024expandable} & 88.34$\pm$0.52& 92.27$\pm$0.53& 82.21$\pm$0.72& 90.76$\pm$0.54& 85.01$\pm$0.69& 89.98$\pm$0.65& 82.10$\pm$0.66& 87.65$\pm$0.46\\ \hline
\textsc{LoRA$^-$DRS} (Ours) & \textbf{89.14}$\pm$0.23& \textbf{92.55}$\pm$0.25 & \textbf{88.69}$\pm$0.15&\textbf{92.25}$\pm$0.24& \textbf{88.39}$\pm$0.33& \textbf{92.02}$\pm$0.19& \textbf{87.29}$\pm$0.31& \textbf{91.29}$\pm$0.29\\ 
\hline
\end{tabular}}
\end{table*}
In this section, we compare \textsc{LoRA$^-$DRS} with other state-of-the-art PEFT methods on CIL benchmarks, testing across various splits and backbone weights. As shown in ~\cref{tab:ming-IN21K}, the accuracy results of different methods on ImageNet-R are presented for different numbers of tasks. 
% Similarly,~\cref{tab:cifar-IN21K} reports results on CIFAR100, and~\cref{tab:domain-IN21K} on DomainNet. 
Similarly,~\cref{tab:cifar-IN21K} reports results on CIFAR100. 
Our method consistently outperforms existing EFCIL methods and other feature drift control methods, such as EASE, InfLoRA, and Adam-NSCL. 
Notably, as the number of tasks increases, the advantage of our method becomes more pronounced, achieving an accuracy margin of 2.58\% to 3.81\% over the runner-up method by the 50th task, highlighting its suitability for continual learning scenarios, where maintaining performance over prolonged task sequences is essential.
% We also present the incremental performance trend for different methods with iBOT-1K in~\cref{fig:iBOT1K} and DINO-1K in~\cref{fig:DINO1K} on ImageNet-R. These results further confirm the robustness of our approach across varying pre-training settings and benchmark splits.

% BWT
To evaluate model stability, we assess the Backward Transfer (BWT) values of different methods on CIFAR-100 and ImageNet-R with 25 tasks. 
As shown in~\cref{tab:bwt}, \textsc{LoRA$^-$DRS} achieves a competitive BWT score of $-3.38$ on CIFAR-100 and $-3.90$ on Imagenet-R, indicating effective feature drift control and consistent performance across tasks. Notably, while methods like InfLoRA and Adam-NSCL also achieve strong BWT scores, they rely on storing statistics from previous tasks to maintain stability, while our method does not. 

\begin{table}[t]
\centering
\caption{Backward transfer (BWT) for different methods on Imagenet-R under 25 incremental tasks.}
\label{tab:bwt}
\begin{tabular}{c|c|c}
\hline
% \hline
\multicolumn{1}{c|}{Method}                                 & CIFAR-100 & \multicolumn{1}{l}{Imagenet-R} \\ \hline
% \multicolumn{1}{c|}{Method}                                  & $BWT$     & $BWT$                           \\ \hline
LoRA-FT & -13.08& -27.52\\ 
LAE~\cite{gao2023unified}  &  -7.06   & -6.32                        \\ 
L2P~\cite{wang2022learning}  & -7.35    & -8.85                         \\ 
DualPrompt~\cite{wang2022dualprompt}    & -7.25    & -6.26                          \\ 
% HiDe-Prompt~\cite{wang2024hierarchical}&   &  \\ 
CODA-Prompt~\cite{smith2023coda}& -5.59  &  -4.24\\ 

InfLoRA~\cite{liang2024inflora}       & -3.70    & -7.56                          \\ 
Adam-NSCL~\cite{wang2021training} &  -4.39 & -3.91                          \\ 
EASE~\cite{zhou2024expandable}& -5.54 & -4.53                          \\ \hline
\textsc{LoRA$^-$DRS} (Ours) &  \textbf{-3.38}& \textbf{-3.90}   \\ 
\hline
% \hline
\end{tabular}
\end{table}

% (Varying the Pre-Trained Model)

\subsection{Detailed Analysis}

\noindent\textbf{Drift evaluation}\quad 
% \cxb{why? what for?}
To further verify that our method effectively controls feature drift, we evaluate it on the ImageNet-R 25 tasks setting and compare feature drift across tasks on previously learned classes. 
% The results are shown~\cref{fig:feat_drifts}, where we present the feature distance of the class center from task $1$ data, extracted using the network at the task $t$, to its corresponding prototypes from the first task. Specifically, we calculate the squared Euclidean distance between the estimated class center and the true class prototype.
\cx{As shown in ~\cref{fig:feat_drifts}, the feature drifts of different methods can be illustrated by measuring the shift in feature representations over tasks.
Specifically, we compute the average squared Euclidean distance between the class center of task $1$ data, as extracted by the network at task $t$, and the corresponding class prototypes from the first task.}
% \lx{The results are shown~\cref{fig:feat_drifts}, where we illustrate the phenomenon of feature drift by measuring the shift in feature representations over tasks. Specifically, we compute the average squared Euclidean distance between the class center of task $1$ data, as extracted by the network at task $t$, and the corresponding class prototype from all classes in the first task.}
% As seen in~\cref{fig:feat_drifts}, 
The distance for the current state-of-the-art PEFT models tends to increase as tasks progress. 
% indicating substantial feature drift in these models under continual learning (CIL) settings. 
Furthermore, InfLoRA attempts to reduce interference, but it still shows persistent feature drift due to reliance on outdated stored task statistics.
% Furthermore, while InfLoRA claiming to mitigate interference from new task learning on old tasks, it suffers persistent feature drift. This occurs because it rely on outdated stored task statistics to filter old task information from the new task space, which limits adaptability in the dynamic CIL environment. 
In contrast, our~\textsc{LoRA$^-$DRS} maintains a steady feature drift curve, demonstrating effective drift control.
% In contrast, our method avoids this dependency by dynamically establishing an appropriate learning space for each new task, effectively controlling feature drift, as evidenced by the steady curve in our model.

\begin{table}[t]
\centering
\caption{The effectiveness of each component in our method on Imagenet-R. \cx{Averaged accuracy ($\overline{ACC}$) is reported.}}
\label{tab:ablation}
\begin{tabular}{cc|ccc}
\hline
% \hline
\multirow{2}{*}{DRS}       & \multirow{2}{*}{ATL}       & \multicolumn{3}{c}{Task}                              \\ \cline{3-5} 
                     &                      & \multicolumn{1}{c|}{10}    & \multicolumn{1}{c|}{25}    & 50    \\ \hline
\ding{55 }                  & \ding{55  }                 & \multicolumn{1}{c|}{76.03} & \multicolumn{1}{c|}{67.13} & 58.56 \\ 
\ding{51 }                  & \ding{55  }                 & \multicolumn{1}{c|}{80.32} & \multicolumn{1}{c|}{78.00} & 76.03 \\ 
\ding{51 }                  & \ding{51  }                 & \multicolumn{1}{c|}{81.52} & \multicolumn{1}{c|}{79.79} & 77.93 \\ 
\hline
% \hline
\end{tabular}
\end{table}
\begin{table}[t]
\centering
\caption{Averaged new class accuracy ($AA_{new}$) on Imagenet-R.}
% \vspace{-0.25cm}
% \caption{Evaluation of ATL in learning new classes $A_{new}$ on Imagenet-R.}
\label{tab:atl_new}
\begin{tabular}{c|ccc}
\hline
% Method & 10 tasks & 25 tasks & 50 tasks \\ \hline
 \multirow{2}{*}{Method} & \multicolumn{3}{c}{Task}                              \\ \cline{2-4}
  & \multicolumn{1}{c|}{10}    & \multicolumn{1}{c|}{25}    & 50    \\ \hline
DRS              & \multicolumn{1}{c|}{79.68} & \multicolumn{1}{c|}{ 77.31} & 75.83 \\ 
\multicolumn{1}{c|}{+ATL}            & \multicolumn{1}{c|}{\textbf{81.23} \footnotesize{(1.55$\uparrow$)}} & \multicolumn{1}{c|}{\textbf{79.18} \footnotesize{(1.87$\uparrow$)}} &  \textbf{77.88} \footnotesize{(2.05$\uparrow$)} \\ 
\hline
\end{tabular}
% \vspace{-0.3cm}
\end{table}

% \subsubsection{Ablation Study}
\noindent\textbf{Ablation Study}\quad
% DRS、w/o DRS(Lora-Fintune)、w/o DRS(null space)
% without triplet loss (w/o L_TL)
In this section, we conduct an ablation study to investigate the effectiveness of each component in~\textsc{LoRA$^-$DRS}, including Drift-Resistant Space (DRS) and Augmented Triplet Loss (ATL). Specifically, we report the incremental performance of different variations on ImageNet-R. 
% We perform experiments to verify the effectiveness of our drift-resistant space (DRS), exploring both training without DRS and training in null space. As shown in~\cref{tab:ablation} , our method consistently outperforms these variants. Specifically, Training without DRS, which relies solely on LoRA for adapting to new tasks, achieves an $\overline{ACC}$ of xxx\%, significantly lower than our method. This result suggests that without intervention, the model suffers from severe forgetting due to feature drift. In the case of Training in null space, we observe a decline in both $\overline{ACC}$ and $ACC$ as the number of tasks increases. This indicates that a static old task feature space fails to effectively capture the dynamic nature of the model's evolution, limiting the model’s stability in preserving old task knowledge. 

As shown in~\cref{tab:ablation}, the performance of training without DRS, relying solely on LoRA for adapting to new tasks, is significantly lower than our method. At the 50-task mark, the model without DRS achieves an $\overline{ACC}$ that is 17.47\% lower than with DRS. This result suggests that, without the DRS intervention, the model experiences severe forgetting due to feature drift.
Additionally, we assess the impact of our ATL on model performance. \cref{tab:ablation} shows that incorporating the ATL in the full \textsc{LoRA$^-$DRS} model leads to an improvement of 1.2\% to 1.9\% in $\overline{ACC}$ compared to the variant without ATL under the same task setting. \lx{\cref{tab:atl_new} \cx{further} show that ATL significantly improves the model's performance \cx{on \emph{new} classes}.}
% in learning new classes.} 
% These indicate that 
Therefore, ATL enhances the model's ability to obtain discriminative features, contributing to better plasticity and accuracy.
\begin{figure}[t]
  \centering
   \includegraphics[width=\linewidth]{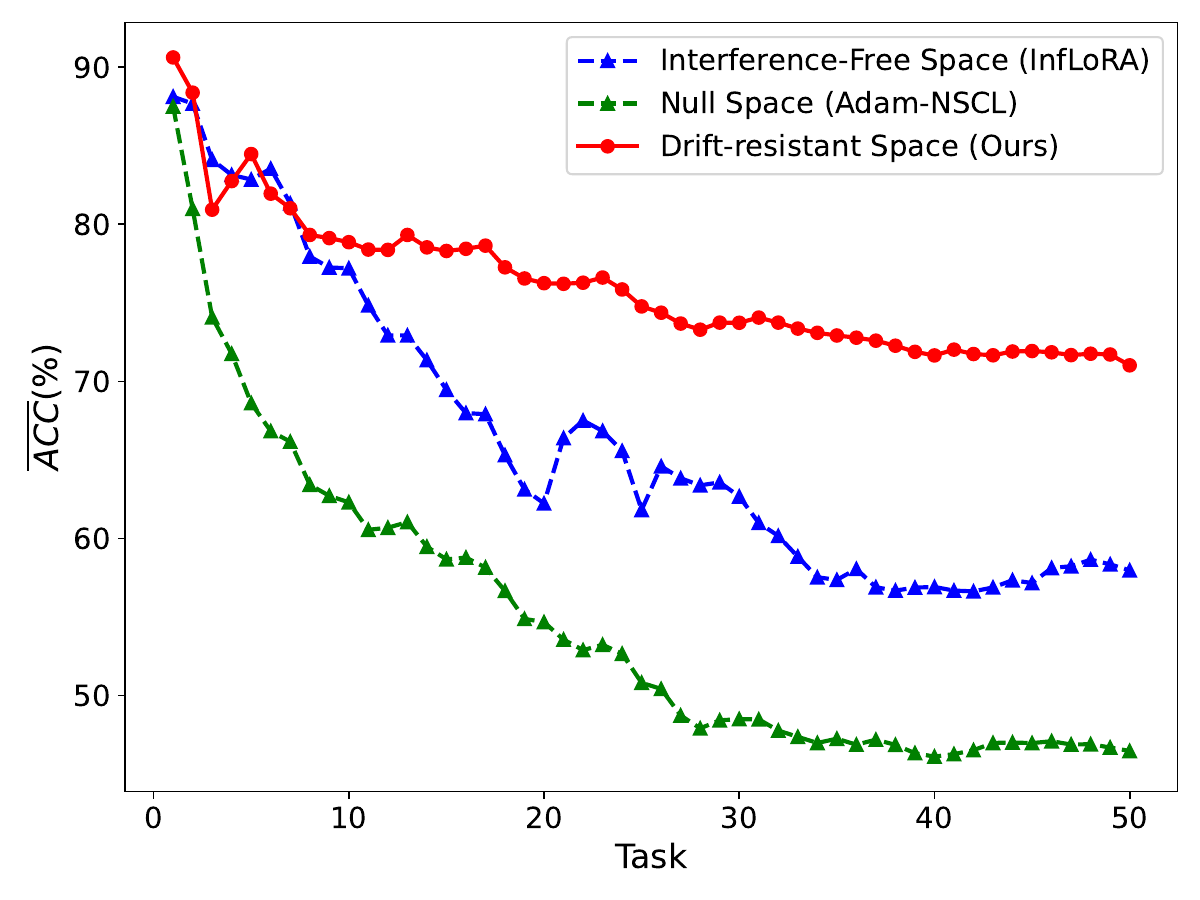}
   \caption{Performance comparison of different space designs on Imagenet-R across 50 incremental tasks.}
    \label{fig:space}
\end{figure}

% \subsubsection{Further Analysis}
\noindent\textbf{Further Analysis}\quad 
% \cxb{specifying them point by point.}
% Analysis of DRS: DRS 选取基向量对应奇异值占比 (85%~100%)?
% Our approach selects a subset of principal components corresponding to the largest singular values to construct the DRS projection matrix based on~\cref{eq:select}. 
% ~\cref{fig:pca_proportion} shows the $\overline{ACC}$ values under different $\varepsilon$ settings on the Imagenet-R dataset. As the results indicate, when $\varepsilon$ is below a certain threshold (xxx), the model's $\overline{ACC}$ is relatively low, suggesting that a significant amount of useful information in DRS is not being retained. In contrast, when $\varepsilon$ exceeds xxx, the $\overline{ACC}$ stabilizes around xxx\%, indicating that the principal components most critical to DRS have been successfully preserved and leveraged for training.
To validate the effectiveness of our DRS, we compared it with two other space designing methods, Interference-Free space (InfLoRA) and Null Space (Adam-NSCL), on ImageNet-R across 50 incremental tasks. As shown in~\cref{fig:space}, which presents the ${ACC}$ values for each task, our DRS demonstrates superior stability and resilience against performance degradation over longer task sequences.
While all methods show some accuracy decline with more tasks, DRS consistently outperforms, maintaining higher accuracy and a more controlled decline. InfLoRA and Adam-NSCL, despite high initial accuracy, experience a marked drop as tasks progress, suggesting their sensitivity to cumulative drift and limitations in preserving old task knowledge. In contrast, our \textsc{LoRA$^-$DRS} effectively mitigates feature drift over time, making it well-suited for long-term incremental learning scenarios.
% The results indicate that, while all three methods show some decline in accuracy as the number of tasks increases, our DRS consistently outperforms other methods. Despite their initial high accuracy, InfLoRA and Adam-NSCL experience a noticeable drop in performance as the tasks progress, suggesting a sensitivity to cumulative drift. This indicates that a static old task feature space fails to effectively capture the dynamic nature of the model's evolution, limiting the model’s stability in preserving old task knowledge.In contrast, our \textsc{LoRA$^-$DRS} maintains a higher accuracy and shows a more gradual and controlled decline, suggesting that it effectively mitigates feature drift over time, making it well-suited for scenarios that require long-term incremental learning.
% 
% Analysis of Triplet loss: other method+L_TL无效因为他们有漂移？

During the training phase, we introduced ATL to enhance class separation, aiming to achieve greater plasticity when learning new classes. However, the effectiveness of ATL relies on effectively controlling feature drift. If drift occurs, the feature space for previous tasks may shift, undermining ATL’s ability to separate new class samples from old ones. 
As shown in~\cref{tab:ATL}, comparing InfLoRA with and without ATL on ImageNet-R reveals a performance drop when ATL is applied, and this drop becomes more pronounced as the number of tasks increases. The $\overline{ACC}$ decreases by 4.81\% after 50 tasks, highlighting that current methods struggle to control feature drift over long task sequences. Without effective drift control, ATL fails to enhance class separation, negatively impacting overall performance.
\begin{table}[t]
\centering
\caption{Results of InfLoRA with ATL on Imagenet-R. The effectiveness of ATL depends on controlling feature drift, as unmanaged drift can undermine model's performance.}
\label{tab:ATL}
\begin{tabular}{c|ccc}
\hline
% \multirow{2}{*}{} & \multicolumn{3}{c}{Task}                                       \\ \cline{2-4} 
\multicolumn{1}{l|}{\rule{0pt}{2.5ex}}                  & \multicolumn{1}{c|}{$\overline{ACC_{10}}$}    & \multicolumn{1}{c|}{$\overline{ACC_{25}}$}    & $\overline{ACC_{50}}$    \\ \hline
InfLoRA           & \multicolumn{1}{c|}{79.93} & \multicolumn{1}{c|}{74.73} & 66.27 \\ 
+ATL              & \multicolumn{1}{c|}{79.85$\downarrow$} & \multicolumn{1}{c|}{71.58$\downarrow$} & 61.46$\downarrow$ \\ 
\hline
\end{tabular}
\end{table}
% 
% \label{tab:ATL}
% \begin{tabular}{c|ccc}
% \hline
% % \hline
% \multirow{2}{*}{} & \multicolumn{3}{c}{Task}                                       \\ \cline{2-4} 
%                  & \multicolumn{1}{c|}{10}    & \multicolumn{1}{c|}{25}    & 50    \\ \hline
% InfLoRA           & \multicolumn{1}{c|}{79.93} & \multicolumn{1}{c|}{74.73} & 66.27 \\ 
% +ATL              & \multicolumn{1}{c|}{79.85$\downarrow$} & \multicolumn{1}{c|}{71.58$\downarrow$} & 61.46$\downarrow$ \\ 
% \hline
% % \hline
% \end{tabular}
% \end{table}
% \noindent\textbf{Memory Usage}\quad

% \lx{Existing methods typically store statistical information from previous tasks to reduce the impact of feature drift. We compare the memory usage of prior methods that explore related ideas. As shown in ~\cref{tab:store_stats}, our method demonstrates the lowest memory requirement with ViT-B/16-IN21K for CIFAR100 50 tasks, as it does not retain statistics from previous tasks. This enables efficient storage and computation while effectively handling feature drifts without explicit feature modeling.}
% \begin{table}[ht]
% \caption{Memory usage of storing statistics on CIFAR100 50-task.}
% % \vspace{-0.25cm}
% \label{tab:store_stats}
% \centering
% \scalebox{0.9}{
% \begin{tabular}{c|ccc}
% \hline
% Method      & Adam-NSCL & InfLoRA & LoRA$^-$ \\ \hline
% Memory (KB) &    720.9       &  2861.3    &   \textbf{0.0}       \\ \hline
% % Memory Size &    720.9       &  2861.3    &   0.0       \\ \hline
% \end{tabular}}
% % \vspace{-0.3cm}
% \end{table}
\section{Conclusion}
\label{sec:conclusion}
% In conclusion, this paper introduces~\textsc{LoRA$^-$DRS} for effective model training in the exemplar-free continual learning (EFCL) setting. 
In conclusion, this paper introduces~\textsc{LoRA$^-$DRS} for effective model training in the exemplar-free continual learning setting. 
Our Drift-Resistant Space (DRS) strategy successfully mitigates feature drift, resulting in enhanced stability and reduced catastrophic forgetting, as demonstrated by consistent performance across long tasks. Through the novel Low-Rank Adaptation Subtraction (LoRA$^-$) approach, DRS minimizes interference from prior tasks without requiring static data storage, allowing the model to focus on new tasks effectively. By enabling Augmented Triplet Loss (ATL) integration to enhance class separation, our method further bolsters learning plasticity. Extensive experiments confirm the state-of-the-art performance of~\textsc{LoRA$^-$DRS} across long task sequences in the EFCIL setting, underscoring its effectiveness and adaptability.

\noindent\textbf{Limitations}\quad 
\cx{While Low-Rank Adaptation (LoRA) modules reduce parameter counts compared to full fine-tuning, their adapter components still introduce incremental model size overhead. Future research could address this limitation by exploring architectures that seamlessly integrate adaptation mechanisms rather than relying on add-on modules.}
% \lx{While LoRA modules are lightweight, they still increase model size due to adapter storage. Future work could explore integrated designs. This method has been validated only on the ViT architecture.}

\noindent\textbf{Acknowledgment}\quad 
This research was supported by the Natural Science Foundation (NSF) for Young Scientists of China (No.62106289).

{
    \small
    \bibliographystyle{ieeenat_fullname}
    \bibliography{main}
}
\clearpage
\setcounter{page}{1}
\maketitlesupplementary

\renewcommand{\thefigure}{\Alph{figure}} 
\renewcommand{\thetable}{\Alph{table}}  
\renewcommand{\thesection}{\Roman{section}}
\setcounter{figure}{0}
\setcounter{table}{0}
\setcounter{section}{0}

% \section{More Experimental Results}

\section{Experiments Across Diverse Datasets}
We apply our method to DomainNet and CUB datasets, following the experimental setup of InfLoRA~\cite{liang2024inflora} and EASE~\cite{zhou2024expandable}, respectively. As shown in Table~\ref{tab:cub}, while our method does not achieve the highest accuracy on DomainNet, it performs comparably to the SOTA methods. DomainNet consists of five short tasks, where our method’s strengths are less evident. However, on CUB, which involves longer tasks, our method excels with an $\overline{ACC}_{20}$ of 92.78\%, outperforming both InfLoRA and EASE. 

\begin{table}[ht]
\centering
\caption{Comparisons in DomainNet and CUB.}
% \vspace{-0.25cm}
\label{tab:cub}
% \scalebox{0.8}{
\begin{tabular}{c|cc|cc}
\hline
\multirow{2}{*}{Method} & \multicolumn{2}{c|}{DomainNet} & \multicolumn{2}{c}{CUB}  \\ 
% \multicolumn{1}{c|}{} & \multicolumn{2}{c|}{CUB} & \multicolumn{2}{c}{DomainNet}  \\ 
% \hline
\cline{2-5}
% \multicolumn{1}{c|}{\rule{0pt}{2.5ex}Method} 
 & \multicolumn{1}{c}{\rule{0pt}{2.5ex}$ACC_{5}$} & \multicolumn{1}{c|}{$\overline{ACC}_{5}$} & \multicolumn{1}{c}{$ACC_{20}$} & \multicolumn{1}{c}{$\overline{ACC}_{20}$} \\ 
% \multicolumn{1}{c|}{\rule{0pt}{2.5ex}Method} & \multicolumn{1}{c}{$ACC_{20}$} & \multicolumn{1}{c|}{$\overline{ACC}_{20}$} & \multicolumn{1}{c}{$ACC_{5}$} & \multicolumn{1}{c}{$\overline{ACC}_{5}$} \\ 
\hline
InfLoRA & 69.68 & \textbf{76.93}& 62.68& 76.57 \\
% Adam-NSCL & & & &  \\
EASE& 66.39 & 72.21& 86.13& 91.68 \\ 
% \hline
Ours & \textbf{70.37} & 76.65 & \textbf{87.74}& \textbf{92.78} \\ 
% \textsc{LoRA$^-$DRS} (Ours) & & & &  \\ 
\hline
\end{tabular}
% }
\vspace{-0.3cm}
\end{table}

\section{Variants for Computing Drift-Resistance Space}
We conduct experiments to evaluate the effectiveness of the LoRA subtraction method. Specifically, we employ the initial pre-trained weights $W_0$ to design DRS.~\cref{tab:w0} presents the results of our method alongside its variant. From the results, we observe that the variant does not perform as effectively as our method, thus LoRA$^-$ is necessary.
% \begin{table}[ht]
%     \centering
%     \caption{DRS Computation with $W_0$ v.s. LoRA$^-$ on CIFAR-100.}
%     \vspace{-0.25cm}
%     \label{tab:w0}
%     \begin{tabular}{c|cc}
%     \hline
% {\rule{0pt}{2.5ex}} {DRS Compute}                             & $\overline{ACC}_{10}$ & $\overline{ACC}_{50}$ \\ \hline
%     % $W_0\to \textsc{DRS}$  
%     $W_0$  
%     &      90.33                 &         78.32              \\ %\hline
%     % $LoRA^- \to \textsc{DRS}$ 
%     % LoRA$^-$ 
%     Ours
%     &    \textbf{92.78}  &                \textbf{  91.29}        \\ \hline
%     \end{tabular}
%     % \vspace{-0.3cm}
% \end{table}

\begin{table}[ht]
    \centering
    \caption{DRS Computation with $W_0$ v.s. LoRA$^-$ on CIFAR-100.}
    % \vspace{-0.25cm}
    \label{tab:w0}
    \scalebox{0.9}{
    \begin{tabular}{c|cc|cc}
    \hline
    % Task

{\rule{0pt}{2.5ex}} %{DRS Compute}                   
& $ACC_{10}$ & $\overline{ACC}_{10}$ & $ACC_{50}$ & $\overline{ACC}_{50}$ \\ \hline
    $W_0\to \textsc{DRS}$  
    % $W_0$  
    & 63.04          &      90.33                 & 52.88          &         78.32              \\ %\hline
    $LoRA^- \to \textsc{DRS}$ 
    % LoRA$^-$ 
    % Ours
    & \textbf{89.40}           &    \textbf{92.78}                    & \textbf{86.82 }        &                \textbf{  91.29}        \\ \hline
    \end{tabular}
    }
    % \vspace{-0.3cm}
\end{table}

\section{Further Performance Analysis}
We provide a detailed analysis of the performance of the old and new classes compared to existing methods~\cite{zhou2024expandable,liang2024inflora,wang2021training,gao2023unified,wang2022learning}, as shown in~\cref{tab:old_new}. Specifically, the results demonstrate a 5.5\% improvement in $A_{old}$ over EASE, and a 9.5\% improvement in $A_{new}$ compared to EASE. Our method is consistently better than all methods on both tasks and thus more stable and plastic. 

\begin{table}[ht]
    \centering
    \caption{The old class accuracy ($A_{old}$) and new class accuracy ($A_{new}$) at different stages of CIFAR-100 50 tasks.}
    % \vspace{-0.25cm}
    % \caption{\cxb{The averaged old accuracy ($A_{old}$) and averaged new accuracy ($A_{new}$) compared on CIFAR-100.}}
    \label{tab:old_new}
    \scalebox{0.7}{
    \begin{tabular}{c|cc|cc|cc|cc}
    % \toprule
    \hline
    & \multicolumn{2}{c}{\cxb{Stage-10}} & \multicolumn{2}{|c}{\cxb{Stage-20}} & \multicolumn{2}{|c}{\cxb{Stage-40}}& \multicolumn{2}{|c}{\cxb{Stage-50}}\\
    % & \multicolumn{2}{c}{20 Tasks} & \multicolumn{2}{|c}{50 Tasks} \\
    % \cmidrule(lr){2-3} \cmidrule(lr){4-5}
    \cline{2-9}
    \textbf{} & \textbf{$A_{old}$} & \textbf{$A_{new}$} & \textbf{$A_{old}$} & \textbf{$A_{new}$}& \textbf{$A_{old}$} & \textbf{$A_{new}$}& \textbf{$A_{old}$} & \textbf{$A_{new}$}\\
    % \midrule
    \hline
    LAE &  90.17  &  92.5   & 83.13&  83.5&  79.22&  80.5&  73.54&  85.5 \\
    L2P &    88.72&     91.5&   79.95&   94.5&  76.33&  90.0&   76.72&  68.5  \\
    % DualPrompt &  87.78&    97.0&   83.39&  80.5 &  79.06&  86.0&  76.29&  79.5 \\
    % CODA-Prompt &    62.68&     82.65&   59.98&   64.3&  55.52&  60.0&  56.08&   53.5 \\
    InfLoRA &  83.83  &  79.0   &  70.76 &  94.0 &  61.29& 67.0 &  60.84&  78.5 \\
    Adam-NSCL &     78.67&    61.5 &   65.24&    79.0&   56.83&   59.5&   53.19&  68.5  \\
    EASE &  92.67   &  95.0   &   87.53 & 94.0  & 81.77& 89.5& 81.34& 76.5 \\
    % \textsc{LoRA$^-$DRS}
    Ours & \textbf{95.61 }  &  \textbf{ 95.0}  & \textbf{  91.34}&  \textbf{94.5 }& \textbf{87.53}& \textbf{92.5}& \textbf{86.84} & \textbf{ 86.0} \\
    % \textsc{LoRA$^-$DRS} (Ours) & \textbf{88.54}  &  87.2  & \textbf{86.84}  & \textbf{ 86.0} \\
    % \bottomrule
    \hline
    \end{tabular}
    }
    % \vspace{-0.3cm}
\end{table}

% \section{Hyperparameter Analysis}
% We perform the hyperparameter analysis for our method. The $\lambda$ parameter in~\cref{eq:loss} is selected through cross-validation on the training set, optimizing for the best performance on the validation set.
% ~\cref{fig:ab_lambda} shows the results of our method with different values of $\lambda$. It can be observed that the performance of $\textsc{LoRA$^-$DRS}$ increases with $\lambda$ initially, and then starts to decrease as $\lambda$ continues to grow.
% \begin{figure}[t]
%   \centering
%   % \fbox{\rule{0pt}{2in} \rule{0.9\linewidth}{0pt}}
%    \includegraphics[width=0.7\linewidth]{CVPR25-LX/images/lambda_cifar.pdf}
%    \caption{Analysis of the hyperparameter $\lambda$ on CIFAR100.}
%     \label{fig:ab_lambda}
% \end{figure}

\section{Memory Usage and Storage Efficiency}
Existing methods typically store statistical information from previous tasks to reduce the impact of feature drift. We compare the memory usage of prior methods that explore related ideas, including InfLoRA and Adam-NSCL. As shown in~\cref{tab:store_stats}, our method demonstrates the lowest memory requirement with ViT-B/16-IN21K for CIFAR100 50 tasks, as it does not retain statistics from previous tasks. Our LoRA$^-$ approach enables efficient storage and computation while effectively handling feature drifts without explicit feature modeling.
\begin{table}[ht]
\caption{Memory usage of storing statistics on CIFAR100 50-task.}
% \vspace{-0.25cm}
\label{tab:store_stats}
\centering
\scalebox{0.95}{
\begin{tabular}{c|ccc}
\hline
Method      & Adam-NSCL & InfLoRA & Ours \\ \hline
Memory (KB) &    720.9       &  2861.3    &   \textbf{0.0}       \\ \hline
% Memory Size &    720.9       &  2861.3    &   0.0       \\ \hline
\end{tabular}}
% \vspace{-0.3cm}
\end{table}

% \section{Rationale}
% \label{sec:rationale}
% % 
% Having the supplementary compiled together with the main paper means that:
% % 
% \begin{itemize}
% \item The supplementary can back-reference sections of the main paper, for example, we can refer to \cref{sec:intro};
% \item The main paper can forward reference sub-sections within the supplementary explicitly (e.g. referring to a particular experiment); 
% \item When submitted to arXiv, the supplementary will already included at the end of the paper.
% \end{itemize}
% % 
% To split the supplementary pages from the main paper, you can use \href{https://support.apple.com/en-ca/guide/preview/prvw11793/mac#:~:text=Delete%20a%20page%20from%20a,or%20choose%20Edit%20%3E%20Delete).}{Preview (on macOS)}, \href{https://www.adobe.com/acrobat/how-to/delete-pages-from-pdf.html#:~:text=Choose%20%E2%80%9CTools%E2%80%9D%20%3E%20%E2%80%9COrganize,or%20pages%20from%20the%20file.}{Adobe Acrobat} (on all OSs), as well as \href{https://superuser.com/questions/517986/is-it-possible-to-delete-some-pages-of-a-pdf-document}{command line tools}.
\end{document}